 \documentclass[pmlr,twocolumn,10pt]{jmlr} 

\jmlrworkshop{Machine Learning for Health (ML4H) 2024} 
\usepackage{bm}

\usepackage{rotating}

\usepackage{booktabs}
\usepackage{siunitx}
\usepackage{titlesec}

\usepackage[switch]{lineno}
\usepackage{natbib}

\theorembodyfont{\upshape}
\theoremheaderfont{\scshape}
\theorempostheader{:}
\theoremsep{\newline}

 \title[Infinite hierarchical contrastive clustering for personal digital envirotyping]{Infinite hierarchical contrastive clustering for personal digital envirotyping}

\author{%
\Name{Ya-Yun Huang}\Email{ya-yun.huang@duke.edu}\\
\addr Duke University, NC, USA
\AND
\Name{Joseph McClernon}\Email{francis.mcclernon@duke.edu}\\
\addr Duke University, NC, USA
\AND
\Name{Jason A. Oliver}\Email{jason-oliver@ouhsc.edu}\\
\addr University of Oklahoma Health Sciences Center, OK, USA
\AND
\Name{Matthew M. Engelhard} \Email{m.engelhard@duke.edu}\\
\addr Duke University, NC, USA
}

\begin{document}

\maketitle

\begin{abstract}
Daily environments have profound influence on our health and behavior. Recent work has shown that \textit{digital envirotyping}, where computer vision is applied to images of daily environments taken during ecological momentary assessment (EMA), can be used to identify meaningful relationships between environmental features and health outcomes of interest. 
To systematically study such effects on an individual level, it is helpful to group images into \textit{distinct} environments encountered in an individual's daily life; these may then be analyzed, further grouped into related environments with similar features, and linked to health outcomes. Here we introduce infinite hierarchical contrastive clustering to address this challenge.
Building on the established contrastive clustering framework, our method a) allows an arbitrary number of clusters without requiring the full Dirichlet Process machinery by placing a stick-breaking prior on predicted cluster probabilities; and b) encourages distinct environments to form well-defined sub-clusters within each cluster of related environments by incorporating a participant-specific prediction loss.
Our experiments show that our model effectively identifies distinct personal environments and groups these environments into meaningful environment types. We then illustrate how the resulting clusters can be linked to various health outcomes, highlighting the potential of our approach to advance the envirotyping paradigm.
\end{abstract}

\paragraph*{Data and Code Availability}

We use two sources of daily environment data in our work. The first source comprises images collected from our study team to evaluate the performance of our method and share the results publicly. The second source includes images and associated health outcomes collected from a larger group of participants through a photo EMA process. This data was used to explore the relationship between daily environments and health outcomes. Due to the lack of written consent from participants for public data sharing, these data cannot be made available for public access. All code needed to define and train our model may be found at \href{https://github.com/engelhard-lab/ih-cc-environtyping}{ih-cc-envirotyping}.

\paragraph*{Institutional Review Board (IRB)}
Recruitment and all study procedures were approved by the Duke University Health System Institutional Review Board, and written consent was obtained from all participants.

\section{Introduction}
\label{sec:intro}

The influence of daily environments on human health and behavior has long been recognized \citep{moser2003environmental}. For example, neighborhood aesthetics and access to green spaces have been linked to depressive moods \citep{rautio2018living,tabet2017neighborhood}, and there are established links between daily environments and obesity, psychosocial health, and more \citep{kirk2010characterizing, amerio2020covid}. Home environments are also important but less studied; for example, personal smoking environments have been shown to elicit the urge to smoke and increase smoking behavior \citep{mcclernon2016hippocampal}. These findings underscore the importance of characterizing individuals' daily environments, which we call \textit{envirotyping}, and analyzing relationships between individual envirotypes and health and behavioral outcomes.

Current technologies such as mobile phones allow us to sample personal daily environments together with other features such as GPS on an ongoing basis at almost no cost and minimal user burden. The resulting images can then be processed via computer vision (CV) models to analyze and predict a variety of health and behavioral outcomes \citep{engelhard2019identifying}. However, rather than connecting environmental images and features directly to outcomes, we would like to identify \textit{distinct} environments that an individual encounters in their daily life, which requires an automated method to accurately group multiple images of the same environment. This would then allow us to analyze each environment, summarize the health outcomes associated with it, and approximate the amount of time the individual spends there. Each of these is important both to understand the influences of environments on health, and to develop environment-aware intervention strategies.

Many existing approaches might be considered for this task, but ultimately fall short. Pre-trained image classification models can identify common environment types, but fail to identify distinct environments or adapt to the broad range of settings encountered in daily life. 
GPS can help differentiate between locations that are far apart, but is not sufficiently precise to distinguish between distinct environments within the home. Most promising are image clustering methods, but to our knowledge they have not been applied to identify distinct personal environments, and consequently are not optimized for this purpose. Here we propose a scalable, effective CV method designed to facilitate digital envirotyping by a) grouping images into distinct environments, and b) grouping multiple distinct environments into clusters with similar environmental features (\textit{i.e.}, environment types).

Our approach, which builds on the contrastive clustering (CC) framework proposed by \citet{li2021contrastive}, is motivated by the observation that the data augmentations applied in contrastive learning mirror the variability encountered during repeated sampling of environmental images. For example, cropping and brightness adjustment simulate changes in vantage point and lighting conditions, respectively. 
However, applying the CC framework directly is not optimal, as it has two key limitations in this context. 

First, CC is limited to a fixed number of clusters, which poses a challenge when the number of distinct environments and environment types is unknown. To address this limitation, we propose an \textit{infinite} contrastive clustering model that does not require us to fix the number of clusters \textit{a priori}.
Whereas the infinite clustering problem has previously been solved by placing a Dirichlet Process (DP) prior on latent features \citep{lee2020neural}, we show that within the contrastive clustering framework, a similar effect can be achieved without the full DP by placing a stick-breaking prior on predicted cluster probabilities.

Second, CC provides only one level of clustering, whereas our goal is not only to identify distinct environments, but also to define groups of environments with similar features.
To address this, we associate clusters with the top level in this hierarchy, \textit{i.e.}, environment types, but incorporate an additional supervisory signal that encourages distinct environments to form well-defined sub-clusters within each environment type.
Specifically, we incorporate a \textit{participant-specific prediction head} (PSH) that predicts the individual to whom each image belongs. Since all images of a given distinct environment belong to the same participant, this indirectly encourages formation of distinct sub-clusters and discourages the model from dividing them between multiple clusters.

To illustrate the use of this method and quantify its effectiveness, we first apply our model to images of daily environments collected by our study team. This allows us to a) verify that images of the same environment are indeed assigned to the same cluster; and b) share the images publicly to illustrate the effectiveness of our approach. We then apply the model to a large photo-augmented EMA dataset designed to study the influence of daily environments on smoking and related outcomes. This provides a case study illustrating how the approach can help us understand relationships between environments and health.

Overall, our goal is to facilitate envirotyping from repeated sampling of images of daily environments. To our knowledge, this is the first study to apply a contrastive learning approach to identify distinct and related environments from a large collection of images. Our specific contributions are as follows:
\begin{itemize}
\item We propose a novel method for digital envirotyping that learns well-defined inter-participant clusters (related environments) and intra-participant sub-clusters (distinct environments).
\item We provide a simple, elegant approach to allow an infinite number of clusters that is tailored to the contrastive clustering framework.
\item We evaluate the proposed method on a dataset of labeled environment images to show that the identified environment clusters are meaningful.
\item We show how the approach can facilitate envirotyping efforts by using it to understand relationships between environments and smoking-related outcomes in a large EMA dataset.
\end{itemize}

\section{Related Work}
Image clustering methods can be divided into those that learn features and clusters sequentially (two-step) versus jointly (end-to-end, \textit{i.e.}, deep clustering).
Notable examples of the former include \citet{mcconville2021n2d}, which applies K-means clustering on a manifold learned by UMAP within the representation space of an Autoencoder; and SCAN \citep{van2020scan}, which learns feature representations through self-supervised pre-training. Deep clustering approaches include \citet{yang2016joint}, which updates clusters and representations jointly within a recurrent framework where clustering occurs in the forward pass, and representation learning in the backward pass. While this and other end-to-end approaches are more effective in learning a manifold where clusters are well-defined, they can suffer from accumulated errors during the alternation between steps. \\
In our work, we build on the contrastive clustering (CC) approach proposed by \citet{li2021contrastive}, which learns instance- and cluster-level representations through additional network layers using contrastive samples. The contrastive learning setting aligns with our goal of clustering environments, where images of the same environment can be viewed as contrastive samples. Variants of this framework include Twin Contrastive Learning (TCL) \citep{li2022twin}, which creates contrastive samples by mixing strong and weak data augmentation and adopts a confidence-based criterion to generate pseudo-labels for fine-tuning; and Graph Contrastive Clustering \citep{zhong2021graph}, which constructs cluster graphs with KNN and applies contrastive learning to them. 

\section{Methods}
\begin{figure*}[h!]
    \vspace{-15pt}  
    \centering
    \includegraphics[width=0.8\textwidth]{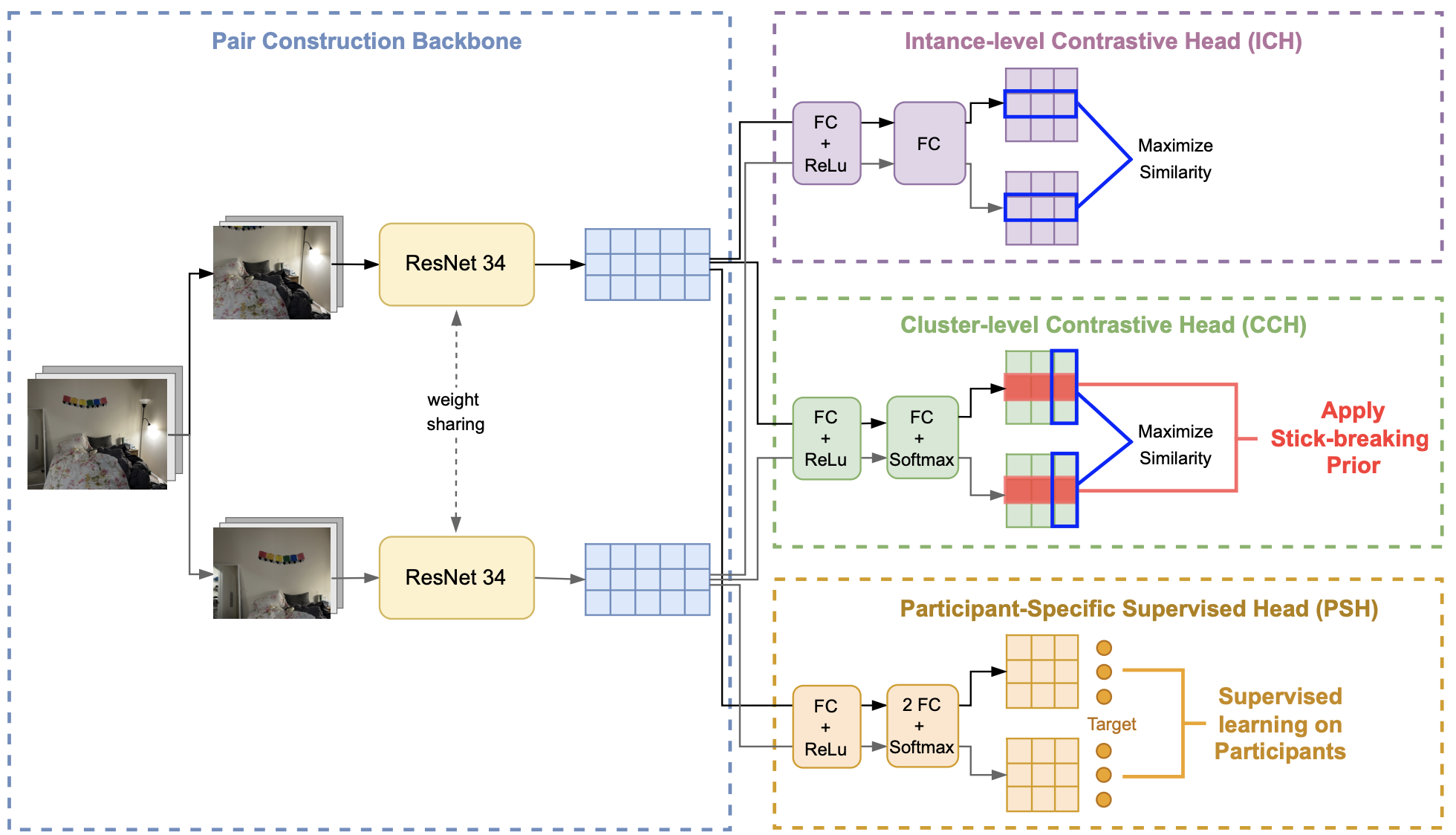}
    \vspace{-10pt}  
    \caption{IH-CC framework. The diagram is adapted and modified from the CC diagram by \citet{li2021contrastive}}.
    \label{fig:diagram}
    \vspace{-28pt}  
\end{figure*}

In this section, we describe the proposed infinite-hierachical contrastive clustering (IH-CC) model in detail. We first describe the CC approach \citep{li2021contrastive}. We then describe how we augment this method to allow an arbitrary number of clusters, and how we incorporate a participant prediction head and loss to encourage the model to learn well-defined sub-clusters representing distinct environments. Finally, we show how all components are integrated together. The diagram of the network is shown in \figureref{fig:diagram}.

\subsection{Base Contrastive Clustering(CC)}
 The base CC model \citep{li2021contrastive} consists of three jointly learned components: a pair construction backbone (PCB), an instance-level contrastive head (ICH), and a cluster-level contrastive head (CCH). First, the PCB applies image augmentation on each sample and constructs data pairs. Data pairs are denoted as positive if they are augmentations of the same image and negative otherwise. Then, an encoder maps these augmented samples to corresponding latent representations. In our work, we chose ResNet34 \citep{he2016deep} as our encoder. Subsequently, the ICH and CCH each take the learned feature matrix as an input to learn instance features and cluster distribution. Both ICH and CCH consist of a fully connected (FC) layer followed by a ReLU activation function and a second FC layer. The output size of ICH is a parameter that can be tuned to effectively learn instance features while that of the CCH should be the desired number of clusters. The CCH applies the SoftMax function to the outputs of the FC layer to generate a cluster distribution for each sample. Then, the ICH and CCH apply a contrastive learning loss to the row and the column space, respectively, in order  to maximize the similarity between positive pairs. These losses are denoted as $L_{ins}$ and $L_{clu}$. Full details may be found in \citet{li2021contrastive}.

\subsection{Stick-breaking(SB) Prior}

To learn the number of clusters to be arbitrarily large and inferred from data rather than fixed \textit{a priori}, we place a stick-breaking (SB) (\textit{i.e.}, Griffiths Engen McCloskey [GEM]) prior on the cluster probabilities $\pi_i(\bm{x})$ assigned by our CCH. 

In the SB construction, we suppose cluster probabilities are generated by an infinite sequence of samples $\{\beta_i\} \sim \text{Beta}(1, \alpha)$, where each $\beta_i$ determines how much of the remaining probability mass not yet assigned to clusters $\{1, ..., i-1\}$ is assigned to cluster $i$. This construction allows us to place a probability distribution on an arbitrary number of cluster probabilities. The mean $\mu=1/(1+\alpha)$ of the aforementioned Beta distribution increases with $\alpha$, so smaller $\alpha$ implies that a greater portion of the remaining mass is placed on each additional cluster. In practice, this results in fewer clusters, since we typically ignore clusters $i>K$, where $K$ is the smallest whole number such that $\sum_{i=1}^K\pi_i > 1-\epsilon$ for some small $\epsilon>0$.

This SB construction is commonly coupled with a base measure $H$ to form a Dirichlet Process; indeed, this approach has been used to cluster latent representations \citep{chapfuwa2020survival} by supposing they are distributed according to an infinite Gaussian mixture model \citep{rasmussen1999infinite}. Here, rather than explicitly modeling the mixture components, we allow these components to be modeled indirectly via the CCH and explicitly model only the cluster (\textit{i.e.}, mixture) probabilities with our SB prior.

To apply this approach to the output cluster probabilities from CCH, we first choose the number of CCH outputs, corresponding to truncating our GEM prior. For each sample, we write $\pi_i(\bm{x})$, where $i$ ranges from $1$ to $K$, to denote the predicted probability that that sample belongs to cluster $i$. The SB construction then implies that a) $\pi_1=\beta_1$, and b) the following relationship holds between the remaining $\pi_i$ and $\beta_i$:

\begin{equation}
\label{eqn:pi_from_beta}
    \pi_i = \beta_i \times \prod_{j=1}^{i-1}(1- \beta_j)
\end{equation}

This implies that we may use the following formula to calculate the $\beta_i$ ($i>1$) from the predicted $\pi_i$:

\begin{equation}
\label{eqn:beta_from_pi}
    \beta_i = \frac{\pi_i}{1- \sum_{i=j}^{i-1} \pi_j}
\end{equation}


Deriving \eqref{eqn:beta_from_pi} from \eqref{eqn:pi_from_beta} is straightforward, and \eqref{eqn:beta_from_pi} is intuitive after noting that the denominator is the portion of the stick remaining after $i-1$ breaks.

Next, we compute the probability density function (PDF) of $\text{Beta}(1,\alpha)$ at each $\beta_i$. By summing the log PDFs across all clusters for a sample, we can evaluate the SB prior for a given predicted cluster distribution and incorporate the corresponding negative log prior $\mathcal{L}_{sb}$ into the loss function with hyperparameter $\lambda_{sb}$ modulating its strength. By tuning the parameters $\alpha$ and $\lambda_{sb}$ of the SB prior, we can guide the model towards more (larger $\alpha$) or fewer (smaller $\alpha$) clusters without explicitly constraining their number. 
We will show this effect in our experiments.

The size of the CCH output, $K$, is chosen to be larger than needed so that the potential number of clusters is effectively infinite. With appropriate choices of $\alpha$ and $\lambda_{sb}$, 
we can effectively place different environments from the same participant in different clusters, yet group similar environments across participants into the same cluster.

\subsection{Participant-Specific Head (PSH)}
A Participant-Specific Head (PSH) is incorporated to encourage the latent representations of images from different participants to be separable, resulting in clearly defined, intra-participant sub-clusters contained within the larger, inter-participant clusters. 
To accomplish this, we add a PSH that predicts the identity of the participant corresponding to each image based on the learned features extracted by the PCB.
The PCH consists of a FC layer followed by a ReLU activation function and another FC layer with Softmax activation, which predicts a probability distribution across participants. A one-hot representation of the participant ID serves as the label to allow calculation of the standard classification (\textit{i.e.}, cross-entropy) loss, denoted $\mathcal{L}_{ps}$.

\subsection{Objective function}

Model learning takes place as a one-stage, end-to-end process in which all parameters in the ICH, CCH, and PSH heads along with those in our CNN-based encoder are jointly optimized. The overall objective is to minimize the following loss:

\begin{equation}
    \mathcal{L} = \mathcal{L}_{ins} + \mathcal{L}_{clu} + \mathcal{L}_{ps} + \lambda_{sb}\mathcal{L}_{sb}
\end{equation}

\section{Data Cohort}

To illustrate the use of our method and quantify its effectiveness, we conduct experiments using data from two different cohorts.  

\subsection{Cohort for image demonstration}
This dataset consists of 484 images of daily environments collected by our 6 study team members. Each was asked to upload at least 10 images for each of the following 6 common living environment types: living room, dining room, kitchen, working area, balcony/porch, and bedroom at various times and from different vantage points. While this dataset is not interesting from a clinical perspective, it is needed to validate our model for two reasons. First, our study team members provided labels indicating where each image was taken (\textit{e.g.}, living room, bedroom), allowing us to verify that images of the same environment are assigned to the same cluster, which is not possible in real EMA datasets. Second, whereas images from the EMA participants may not be shared to protect privacy, these data may be shared publicly to illustrate the effectiveness of our approach.

\subsection{Cohort from PhotoEMA collection}
The second dataset is a large photo-augmented EMA (photoEMA) dataset which we use to illustrate how our approach can be applied to study relationships between environments and health. This dataset consists of 8008 images from 77 active smokers. Participants were prompted multiple times per day over a two-week study period to complete photoEMA, which includes taking a picture of their immediate surrounding environment and answering questionnaire items assessing current smoking, urge to smoke, mood, stress, sadness, restlessness, and fatigue. Most of these items were assessed using rating scales, but all were binarized for the present analysis.

\section{Results on Convenience Sample}

In this section, we apply our approach to images taken by members of our study team to assess the effectiveness of the proposed approach. We demonstrate that: (1) the SB prior effectively guides the model output towards a desired number of clusters; (2) our model successfully identifies distinct environments on an individual level; and (3) our model captures similar environments across participants.

\subsection{Effects of the stick-breaking prior}

\begin{table}[htbp]
  \centering
  \label{tab:hyperparams_clusters}
  \begin{tabular}{|l|c|}
    \hline
    \textbf{SB Prior} & \textbf{Num Clusters} \\ \hline
    $\alpha=1.5,\:\: \lambda =0 $ & $40$\\ \hline
    $\alpha=1.5,\:\:\lambda =1$ & $20$\\ \hline
    $\alpha=2.0,\:\: \lambda =1 $ & $22$\\ \hline    
    $\alpha=3.0,\:\:\lambda =1  $ & $26$\\ \hline
    $\alpha=5.0,\:\:\lambda =1 $ & $34$\\ \hline

  \end{tabular}
  \caption{Effects of SB prior on number of clusters}
  \label{1}
\end{table}

Here, we show that different SB priors result in varying numbers of clusters (see \tableref{1}). In this experiment, the CCH size is fixed at 40, and the strength ($\lambda_{sb}$) of the SB prior is set to 1. First, we find that when the SB prior is turned off by setting $\lambda$ to $0$, the final number of clusters matches the size of the CCH. Furthermore, different choices of the $\text{Beta}$ distribution, achieved by adjusting the $\alpha$ parameter for $\text{Beta}(1, \alpha)$, result in varying numbers of clusters. A smaller $\alpha$ value led to fewer clusters.

\subsection{Cluster Evaluation}

Our study team's images consist of 29 distinct environments belonging to 6 well-defined environment types. We hypothesized that between 10 and 20 clusters would be sufficient to distinguish distinct environments while clustering similar environments together. Thus, we set our $\alpha$ for the SB prior to $\alpha=1.5$. Given that the mean of $\text{Beta}(1, 1.5)$ is $0.4$, this prior is maximized when approximately $99\%$ of the stick is allocated to the first $10$ stick breaks. We set $\lambda_{sb}$ to 1 and the size of the CCH to $40$.

\paragraph{Cluster Labeling}
The IH-CC model produced $20$ clusters. For each participant, each of the associated clusters was manually assigned a label corresponding to one of the elicited environment types. If the cluster consisted of images from multiple environment types, it was labeled as ``Miscellaneous". Output clusters from other models for comparison were also labeled under this rule.

\paragraph{Evaluation Metric}
The effectiveness of the approach was quantified using the Normalized Mutual Information (NMI) and accuracy (ACC). We define predictions as correct whenever an image from an environment $A$ is assigned to a cluster with label $A$.

\paragraph{Models for comparison}

To evaluate our IH-CC model's effectiveness, we compared it with three baseline models. Two of these baselines, the CC model \citep{li2021contrastive} and the Twin-Contrastive-Learning (TCL) model \citep{li2022twin}, are contrastive learning based, allowing us to isolate the effect of our proposed approach compared to the most closely related alternatives. We also included a deep clustering (DC) model \citep{caron2018deep} to highlight the role of contrastive learning in our envirotyping task. These models require a fixed number of clusters; the number was set to 20, which was the number of clusters generated by IH-CC.

\subsubsection{Individual cluster evaluation}

\begin{figure}[h]
\vspace{-5pt}
\centering
  \includegraphics[width=0.45\textwidth]{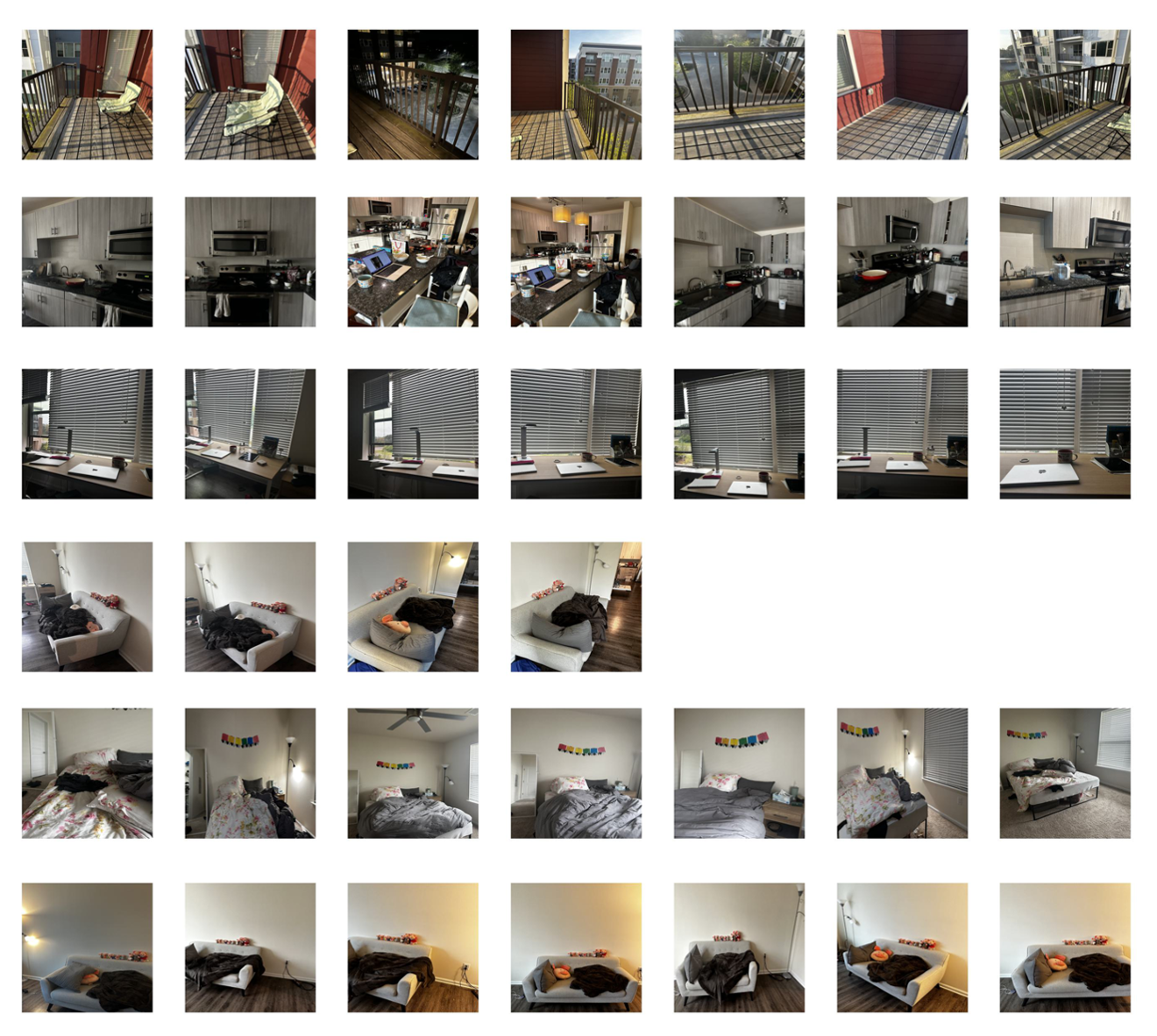}
  \vspace{-5pt}
  \caption{Individual Clusters for P001}
  \label{fig:Individual Clusters}
\end{figure}

Our results show that IH-CC is capable of identifying distinct environments captured by individual participants. \figureref{fig:Individual Clusters} shows clusters for a single participant with each row representing a cluster. We can see that most images of the same environment are assigned to the same cluster, indicating that the model successfully clusters different images of the same environment. In a few cases, images of the same environment are divided between two clusters (see \textit{e.g.} rows 4 and 6 in \figureref{fig:Individual Clusters}). According to the result shown in \tableref{tab:performance}, the NMI and ACC of IH-CC models are both higher than those of the three baseline models. Also, the performances of all CC models are substantially higher than that of DC, indicating the effectiveness of CC framework in the envirotyping setting. Our model assigned an average of $8.3$ clusters per participant, compared to $11.5$ with the CC model. Given that each participant took pictures of at most 6 environments, these results provide initial evidence that the proposed method improves upon CC when clustering similar environments.

\begin{table}[htbp]
    \centering
    \begin{tabular}{l c c}
        \hline
        & NMI & ACC \\ \hline
        IH-CC with PS & 0.516 & 0.791 \\
        IH-CC without PS & 0.550 & 0.782 \\
        CC & 0.496 & 0.754 \\
        TCL & 0.511 & 0.323 \\
        DC & 0.074 & 0.314 \\ \hline
    \end{tabular}
    \caption{Clustering performance by method}
    \label{tab:performance}
    \vspace{-15pt}
\end{table}

\subsubsection{Cross-participant cluster evaluation}
\begin{figure*}[h]
  \centering
  \includegraphics[width=0.8\textwidth]{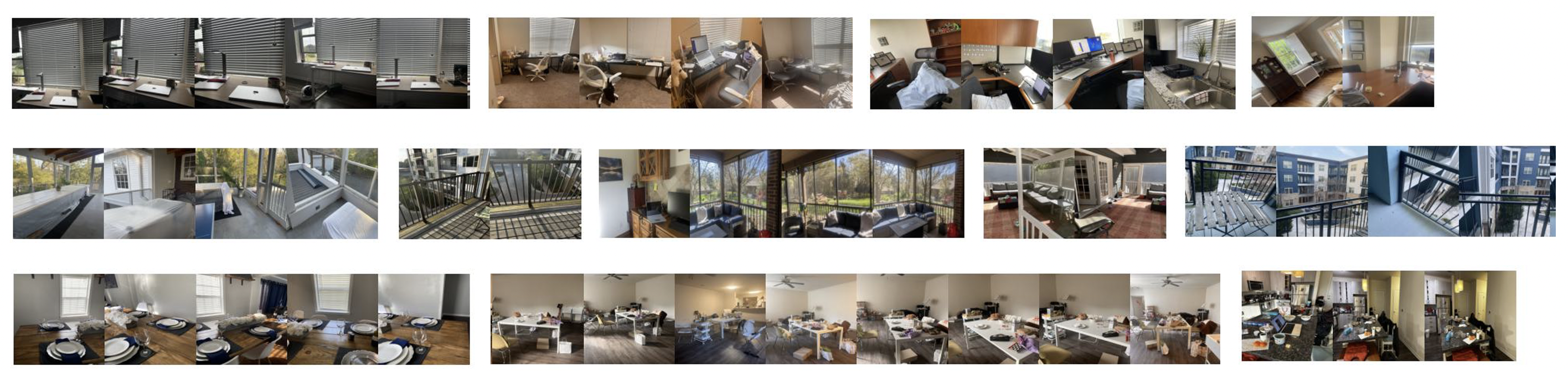}
  \caption{Example of inter-participant clusters (rows) and intra-participant clusters (divided within rows)}
  \label{fig:Cross Participant Clusters}
  \vspace{-10pt}  
\end{figure*}
Our model also produces clusters of environment types across participants. With the SB prior, we can encourage the model to output fewer clusters so that clusters are shared across participants.  We showcase $3$ clusters in Figure ~\ref{fig:Cross Participant Clusters} as examples. Each row contains images from the same cluster, and each group of images are from the same participant.
Row 1 contains images of the ``Working area" environment taken by 4 participants. The cluster captures similar items, including desks, desk chairs, and computer screens, across participants' images. Row 2 contains images of ``Porch" environments, whereas row 3 contains images of ``Dining room" environments.

\titlespacing*{\section}{0pt}{5pt}{*0}
\section{Results on PhotoEMA Study of Smokers' Daily Environments}

We linked the clustering of PhotoEMA images to various health outcomes. Using images from 50 participants with a SB prior of $\alpha=10$, 
we obtain 83 clusters of environment types. Our analysis focused on 66 clusters with $>$10 images, excluding underrepresented clusters. We averaged the outcome scores for each cluster and present the top- and bottom-5 clusters ranked by environment-specific smoking rates in \tableref{tab:clinical_outcome}, with cluster labels assigned manually.

Results indicate high smoking rates in bathrooms, green areas, and porches, while lower rates are observed in bedrooms, kitchens, and other indoor spaces. Clusters with the highest rates of cigarette craving exclusively depict outdoor settings. Feelings of stress, sadness, and fatigue are most common in clusters featuring dark environments, and restlessness is linked to clusters with bathroom images.

\begin{figure}[h]
\centering
  \includegraphics[width=0.4\textwidth]{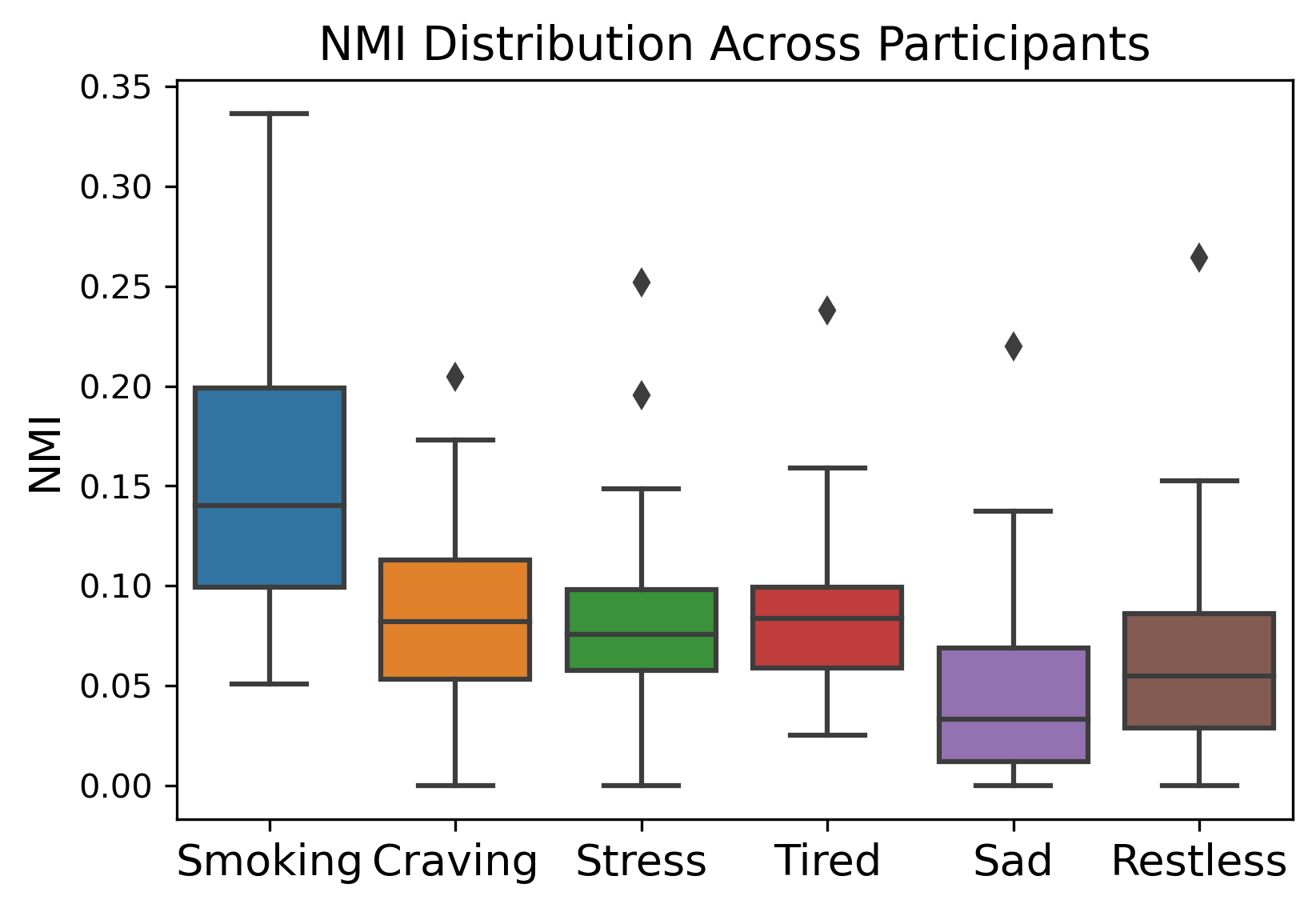}
  \caption{NMI across participants}
  \vspace{-8pt}
  \label{fig:participant NMI}
\end{figure}

Beyond analyzing associations between environment types (\textit{i.e.}, clusters) and outcomes, we aim to understand inter-participant differences in the degree to which environmental factors influence their behaviors and outcomes. To quantify this, we calculate the NMI between cluster membership and each of several behaviors and outcomes for each participant. Intuitively, the NMI measures the degree to which a particular outcome is environmentally linked in a given participant. \figureref{fig:participant NMI} shows that smoking is strongly tied to specific environments; tired and stress are somewhat environmentally linked; and sad and restless are less environmentally linked.

\begin{table*}[h]
\centering

\begin{tabular}{@{}lcccccc@{}}
\toprule
Cluster Label                   & Smoking & Craving & Stress & Tired & Sad & Restless \\ \midrule
Bathroom                        & 0.933   & 0.200   & 0.400  & 1.200 & 0.200 & 0.333    \\
Green Area in the Daytime 1     & 0.910   & 0.469   & 0.455  & 0.593 & 0.138 & 0.414    \\
Porch in the Daytime            & 0.892   & 0.464   & 0.571  & 0.982 & 0.267 & 0.875    \\
Green Area in the Daytime 2     & 0.882   & 0.414   & 0.514  & 0.595 & 0.180 & 0.459    \\
Green Area in the Night         & 0.862   & 0.517   & 0.414  & 1.276 & 0.310 & 0.586    \\
...                             & ......   &  ...... & ......  & ...... & ......  &  ......   \\
Indoor with a view of door                  & 0.429   & 0.321   & 0.786  & 1.176 & 0.143 & 0.357    \\
Bedroom                         & 0.426   & 0.204   & 0.704  & 1.141 & 0.148 & 0.519    \\
Kitchen                         & 0.425   & 0.275   & 0.300  & 0.800 & 0.135 & 0.375    \\
Miscellaneous                   & 0.400   & 0.200   & 0.533  & 0.867 & 0.000 & 0.800    \\
Miscellaneous                   & 0.250   & 0.300   & 0.350  & 0.700 & 0.050 & 0.450    \\ \bottomrule
\end{tabular}
\caption{Outcome rate in each cluster, sorted by the smoking rate}
\label{tab:clinical_outcome}
\vspace{-10pt}  
\end{table*}

\titlespacing*{\section}{0pt}{11pt}{*0}
\section{Ablation Study on PSH}

\begin{figure}[h]
\centering
  \includegraphics[width=0.45\textwidth]
  {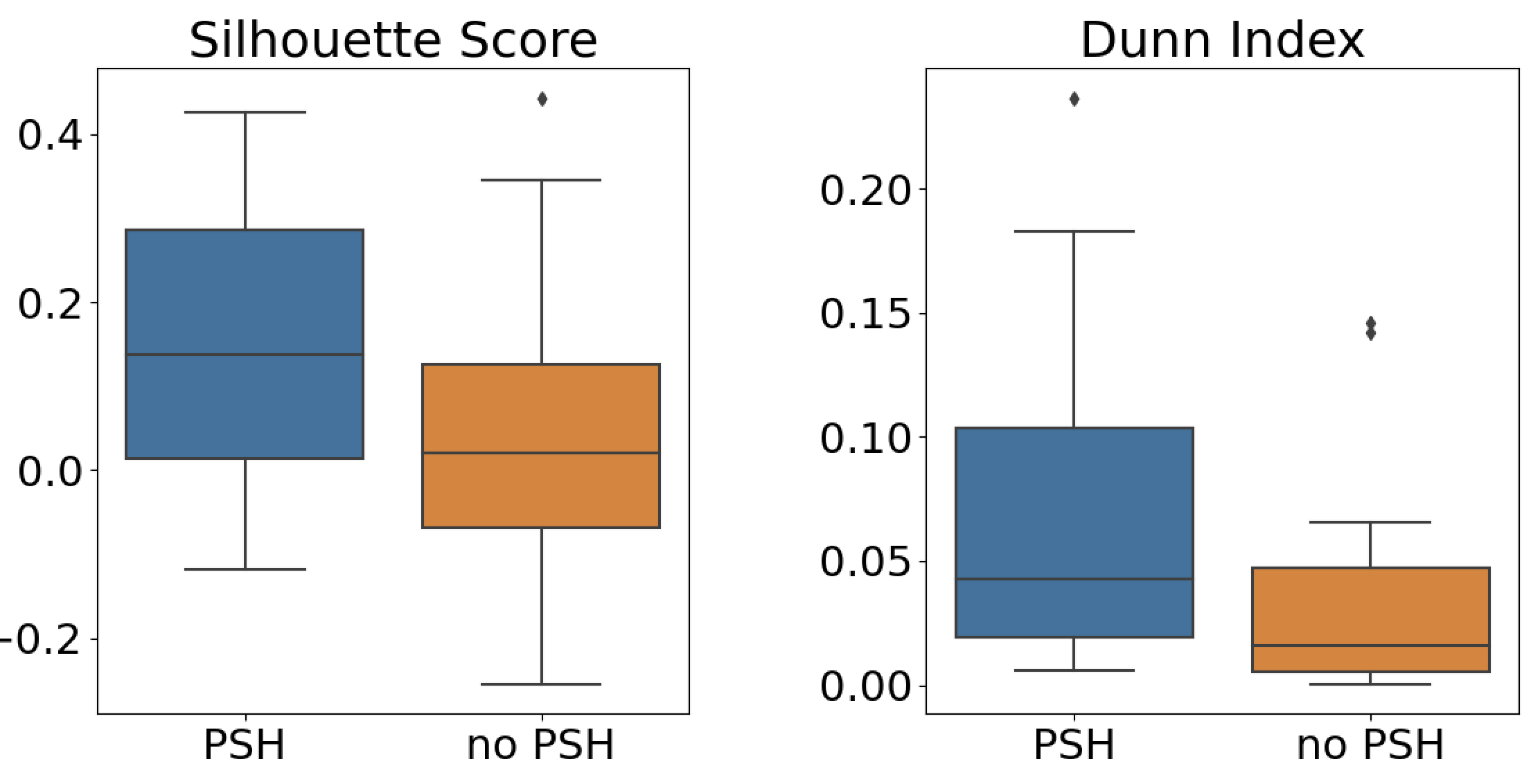}
  \vspace{-5pt}
  \caption{Including the PSH increases the silhouette score (left) and Dunn index (right) of participant-specific subclusters, indicating tighter clustering of distinct environments.}
  \label{fig:box}
  \vspace{-20pt}  
\end{figure}

To evaluate the impact of PSH on grouping intra-participant sub-clusters within larger inter-participant clusters, we conducted ablation studies using the photoEMA dataset, since the study team dataset is not large or varied enough for sub-clustering to be a challenging task.
In the photoEMA dataset, we calculated the \textit{Silhouette Score} and \textit{Dunn Index} to evaluate the compactness of participant-specific clusters -- which represent distinct physical environments -- and their separation from other participants' clusters within each environment type cluster. \figureref{fig:box} shows that the clusters from the IH-CC model with PSH have higher Silhouette Scores and Dunn Index values compared to the IH-CC model without PSH. These results provide quantitative evidence that including the PSH yields more well-defined participant-specific clusters.
\section{Discussion}

In this study, we introduced a novel method for clustering distinct environments present in photoEMA images. Our approach, IH-CC, builds on the CC model \citep{li2021contrastive} to facilitate study of relationships between environments and health outcomes and behaviors. We improve upon existing methods by a) allowing the number of clusters to be inferred from data rather than fixed \textit{a priori} by applying a SB prior to the predicted cluster probabilities; and b) incorporating a supervised Participant-Specific Head that encourages images from different participants to occupy distinct regions of feature space, leading to more clearly defined intra-participant sub-clusters and higher quality inter-participant clusters. Results show that our model groups images into \textit{distinct personal environments}; these are in turn clustered into \textit{environment types} with similar features.

Our real-world envirotyping case study illustrates how IH-CC can facilitate study of relationships between environments and health. Two factors likely contribute to its ability to outperform existing CC models and the DC model. First, our PSH provides a supervisory signal uniquely suited for envirotyping studies, in which we have multiple images from each participant. To our knowledge, repeated measures have not been considered previously in a CC setting, but apply to the many healthcare settings in which there are multiple images per participant. Second, the flexibility provided by our SB prior may help in learning a more appropriate cluster structure compared to other approaches. Additionally, we hypothesize that contrastive methods outperform DC due to the strong parallel between typical image augmentation procedures, which are central to the contrastive learning approach, and the repeated sampling procedure used to acquire images of everyday life across both of our datasets.

\paragraph{Limitations}
This study has important limitations. Although our model was able to capture similarities across environments, its definition of similarity may not align with human perception. For instance, while the model might identify similarity based on color or clothing patterns, humans might perceive similarity based on shared furniture types or layouts. Additionally, while our model is more flexible than those that fix the number of clusters, it still requires tuning the SB prior to ensure clusters are sufficiently broad or granular, respectively, for a given application. As the number of participants increases, it becomes challenging to inspect each cluster, which may lead to ambiguity in determining an optimal configuration.

\paragraph{Future Work}
Identifying distinct personal environments from repeated image capture is key to envirotyping efforts, as it allows us to study relationships between a person's daily environments and their health and behavior.

Individual-level analyses like the ones illustrated here are helpful to provide tailored guidance to support behavior change or structure environment-aware interventions. Recognizing environmentally-linked behaviors is a first step toward developing a behavior change strategy. On the group level, the model can identify environmentally-linked outcomes, facilitate further study of relevant environmental factors, and quantify the differential effectiveness of behavioral interventions between environments. While our study focuses on smoking-related outcomes, the envirotyping paradigm has broad health relevance. Future studies will focus on environmental determinants of mental health and other addictive behaviors.

In conclusion, the importance of envirotyping in health is well-known, but recent advances in mobile technology and computer vision now make personal envirotyping feasible. This work provides a novel, scalable, and effective method to cluster personal environments, in turn facilitating greater study and understanding of how daily environments influence human health and behavior.

\bibliography{jmlr-sample}

\begin{thebibliography}{18}
\providecommand{\natexlab}[1]{#1}
\providecommand{\url}[1]{\texttt{#1}}
\expandafter\ifx\csname urlstyle\endcsname\relax
  \providecommand{\doi}[1]{doi: #1}\else
  \providecommand{\doi}{doi: \begingroup \urlstyle{rm}\Url}\fi

\bibitem[Amerio et~al.(2020)Amerio, Brambilla, Morganti, Aguglia, Bianchi,
  Santi, Costantini, Odone, Costanza, Signorelli, et~al.]{amerio2020covid}
Andrea Amerio, Andrea Brambilla, Alessandro Morganti, Andrea Aguglia, Davide
  Bianchi, Francesca Santi, Luigi Costantini, Anna Odone, Alessandra Costanza,
  Carlo Signorelli, et~al.
\newblock Covid-19 lockdown: housing built environment’s effects on mental
  health.
\newblock \emph{International journal of environmental research and public
  health}, 17\penalty0 (16):\penalty0 5973, 2020.

\bibitem[Caron et~al.(2018)Caron, Bojanowski, Joulin, and Douze]{caron2018deep}
Mathilde Caron, Piotr Bojanowski, Armand Joulin, and Matthijs Douze.
\newblock Deep clustering for unsupervised learning of visual features.
\newblock In \emph{Proceedings of the European conference on computer vision
  (ECCV)}, pages 132--149, 2018.

\bibitem[Chapfuwa et~al.(2020)Chapfuwa, Li, Mehta, Carin, and
  Henao]{chapfuwa2020survival}
Paidamoyo Chapfuwa, Chunyuan Li, Nikhil Mehta, Lawrence Carin, and Ricardo
  Henao.
\newblock Survival cluster analysis.
\newblock In \emph{Proceedings of the ACM Conference on Health, Inference, and
  Learning}, pages 60--68, 2020.

\bibitem[Engelhard et~al.(2019)Engelhard, Oliver, Henao, Hallyburton, Carin,
  Conklin, and McClernon]{engelhard2019identifying}
Matthew~M Engelhard, Jason~A Oliver, Ricardo Henao, Matt Hallyburton,
  Lawrence~E Carin, Cynthia Conklin, and F~Joseph McClernon.
\newblock Identifying smoking environments from images of daily life with deep
  learning.
\newblock \emph{JAMA Network Open}, 2\penalty0 (8):\penalty0 e197939--e197939,
  2019.

\bibitem[He et~al.(2016)He, Zhang, Ren, and Sun]{he2016deep}
Kaiming He, Xiangyu Zhang, Shaoqing Ren, and Jian Sun.
\newblock Deep residual learning for image recognition.
\newblock In \emph{Proceedings of the IEEE conference on computer vision and
  pattern recognition}, pages 770--778, 2016.

\bibitem[Kirk et~al.(2010)Kirk, Penney, and McHugh]{kirk2010characterizing}
Sara~FL Kirk, Tarra~L Penney, and T-LF McHugh.
\newblock Characterizing the obesogenic environment: the state of the evidence
  with directions for future research.
\newblock \emph{Obesity reviews}, 11\penalty0 (2):\penalty0 109--117, 2010.

\bibitem[Lee et~al.(2020)Lee, Ha, Zhang, and Kim]{lee2020neural}
Soochan Lee, Junsoo Ha, Dongsu Zhang, and Gunhee Kim.
\newblock A neural dirichlet process mixture model for task-free continual
  learning.
\newblock \emph{arXiv preprint arXiv:2001.00689}, 2020.

\bibitem[Li et~al.(2021)Li, Hu, Liu, Peng, Zhou, and Peng]{li2021contrastive}
Yunfan Li, Peng Hu, Zitao Liu, Dezhong Peng, Joey~Tianyi Zhou, and Xi~Peng.
\newblock Contrastive clustering.
\newblock In \emph{Proceedings of the AAAI conference on artificial
  intelligence}, volume~35, pages 8547--8555, 2021.

\bibitem[Li et~al.(2022)Li, Yang, Peng, Li, Huang, and Peng]{li2022twin}
Yunfan Li, Mouxing Yang, Dezhong Peng, Taihao Li, Jiantao Huang, and Xi~Peng.
\newblock Twin contrastive learning for online clustering.
\newblock \emph{International Journal of Computer Vision}, 130\penalty0
  (9):\penalty0 2205--2221, 2022.

\bibitem[McClernon et~al.(2016)McClernon, Conklin, Kozink, Adcock, Sweitzer,
  Addicott, Chou, Chen, Hallyburton, and DeVito]{mcclernon2016hippocampal}
F~Joseph McClernon, Cynthia~A Conklin, Rachel~V Kozink, R~Alison Adcock,
  Maggie~M Sweitzer, Merideth~A Addicott, Ying-hui Chou, Nan-kuei Chen,
  Matthew~B Hallyburton, and Anthony~M DeVito.
\newblock Hippocampal and insular response to smoking-related environments:
  neuroimaging evidence for drug-context effects in nicotine dependence.
\newblock \emph{Neuropsychopharmacology}, 41\penalty0 (3):\penalty0 877--885,
  2016.

\bibitem[McConville et~al.(2021)McConville, Santos-Rodriguez, Piechocki, and
  Craddock]{mcconville2021n2d}
Ryan McConville, Raul Santos-Rodriguez, Robert~J Piechocki, and Ian Craddock.
\newblock N2d:(not too) deep clustering via clustering the local manifold of an
  autoencoded embedding.
\newblock In \emph{2020 25th international conference on pattern recognition
  (ICPR)}, pages 5145--5152. IEEE, 2021.

\bibitem[Moser and Uzzell(2003)]{moser2003environmental}
Gabriel Moser and David Uzzell.
\newblock Environmental psychology.
\newblock \emph{Comprehensive handbook of psychology}, 5:\penalty0 419--445,
  2003.

\bibitem[Rasmussen(1999)]{rasmussen1999infinite}
Carl Rasmussen.
\newblock The infinite gaussian mixture model.
\newblock \emph{Advances in neural information processing systems}, 12, 1999.

\bibitem[Rautio et~al.(2018)Rautio, Filatova, Lehtiniemi, and
  Miettunen]{rautio2018living}
Nina Rautio, Svetlana Filatova, Heli Lehtiniemi, and Jouko Miettunen.
\newblock Living environment and its relationship to depressive mood: A
  systematic review.
\newblock \emph{International journal of social psychiatry}, 64\penalty0
  (1):\penalty0 92--103, 2018.

\bibitem[Tabet et~al.(2017)Tabet, Sanders, Schootman, Chang, Wolinsky,
  Malmstrom, and Miller]{tabet2017neighborhood}
Maya Tabet, Erin~A Sanders, Mario Schootman, Jen~Jen Chang, Fredric~D Wolinsky,
  Theodore~K Malmstrom, and Douglas~K Miller.
\newblock Neighborhood conditions and psychosocial outcomes among middle-aged
  african americans: A cross-sectional analysis.
\newblock \emph{Journal of primary care \& community health}, 8\penalty0
  (2):\penalty0 63--70, 2017.

\bibitem[Van~Gansbeke et~al.(2020)Van~Gansbeke, Vandenhende, Georgoulis,
  Proesmans, and Van~Gool]{van2020scan}
Wouter Van~Gansbeke, Simon Vandenhende, Stamatios Georgoulis, Marc Proesmans,
  and Luc Van~Gool.
\newblock Scan: Learning to classify images without labels.
\newblock In \emph{European conference on computer vision}, pages 268--285.
  Springer, 2020.

\bibitem[Yang et~al.(2016)Yang, Parikh, and Batra]{yang2016joint}
Jianwei Yang, Devi Parikh, and Dhruv Batra.
\newblock Joint unsupervised learning of deep representations and image
  clusters.
\newblock In \emph{Proceedings of the IEEE conference on computer vision and
  pattern recognition}, pages 5147--5156, 2016.

\bibitem[Zhong et~al.(2021)Zhong, Wu, Chen, Huang, Deng, Nie, Lin, and
  Hua]{zhong2021graph}
Huasong Zhong, Jianlong Wu, Chong Chen, Jianqiang Huang, Minghua Deng, Liqiang
  Nie, Zhouchen Lin, and Xian-Sheng Hua.
\newblock Graph contrastive clustering.
\newblock In \emph{Proceedings of the IEEE/CVF international conference on
  computer vision}, pages 9224--9233, 2021.

\end{thebibliography}

\end{document}